\documentclass{bmvc2k}
\usepackage{graphicx}
\usepackage{amssymb}
\pdfoutput=1
\usepackage{multirow}
\usepackage[percent]{overpic}
\usepackage{caption,booktabs}
\usepackage[export]{adjustbox}
\usepackage{scalerel}

\DeclareMathOperator*{\argmin}{arg\,min}

\title{Content and Colour Distillation for Learning Image Translations with the Spatial Profile Loss}

\addauthor{M. Saquib Sarfraz\textsuperscript{1,}}{saquib.sarfraz@kit.edu}{2}
\addauthor{Constantin Seibold}{constantin.seibold@kit.edu}{1}
\addauthor{Haroon Khalid}{haroon.khalid49@gmail.com}{1}
\addauthor{Rainer Stiefelhagen}{cvhci.anthropomatik.kit.edu}{1}

\addinstitution{
Institute of Anthropomatics \& Robotics\\
Karlsruhe Institute of Technology\\
Germany
 }
 
\addinstitution{
Autonomous Systems\\
Daimler TSS\\
Germany
}

\runninghead{Sarfraz et al.}{Image Translations with the Spatial Profile loss}

\def\etal{\emph{et al}\bmvaOneDot}

\begin{document}

\maketitle

\newcommand*{\rom}[1]{\uppercase\expandafter{\romannumeral #1\relax}}

\begin{center}
\setlength{\abovecaptionskip}{0pt}
\setlength{\belowcaptionskip}{0pt}
\vspace{-0.5cm}
\includegraphics[width=\linewidth]{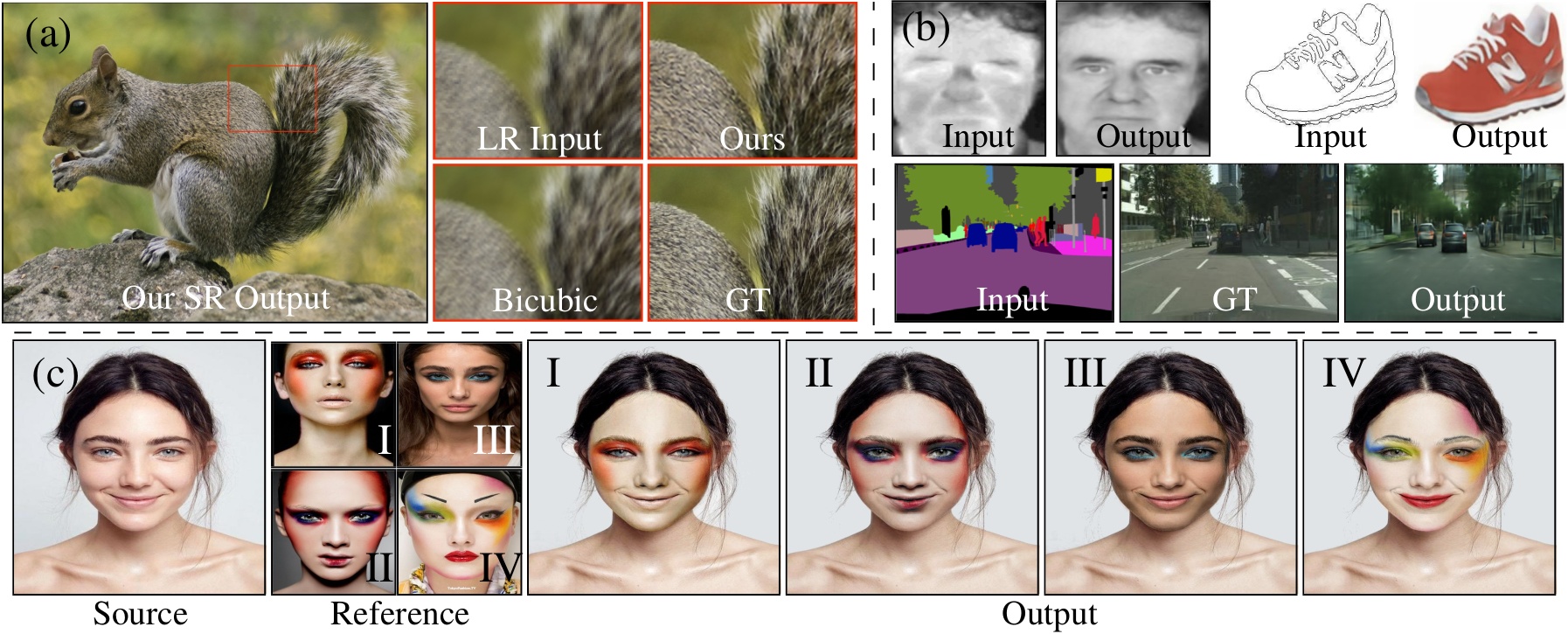}
\captionof{figure}{Various results synthesized by training single generator with our proposed loss formulation. a). Single Image super resolution b). Image-to-image domain mappings and c). Photo realistic reference makeup transfer: We can generate a HD $2048 \times 2048$ portrait with transferred makeup for a given low resolution $512 \times 512$ source and reference pair.}
\label{fig:intro}

\end{center}

\begin{abstract}

Generative adversarial networks has emerged as a defacto standard for image translation problems. To successfully drive such models, one has to rely on additional networks e.g., discriminators and/or perceptual networks. Training these networks with pixel based losses alone are generally not sufficient to learn the target distribution. In this paper, we propose a novel method of computing the loss directly between the source and target images that enable proper distillation of shape/content and colour/style. We show that this is useful in typical image-to-image translations allowing us to successfully drive the generator without relying on additional networks. We demonstrate this on many difficult image translation problems such as image-to-image domain mapping, single image super-resolution and photo realistic makeup transfer. Our extensive evaluation shows the effectiveness of the proposed formulation and its ability to synthesize realistic images.

\end{abstract}
\section{Introduction}
\label{sec:intro}
\vspace{-4pt}

Generative Adversarial Networks (GANs) by Goodfellow et al. \cite{GAN} have dominated the image translation applications~\cite{pix2pix, cyclegan, DualGAN,stargan,liunvidia,Zhunips}. Adversarial training is challenging as it requires two networks to compete with each other.
No matter how attractive the idea of pitting two networks against each other sound, studies have demonstrated the difficulty attached to the process of adversarial training for finding the saddle point that is highly susceptible to hyperparameter settings~\cite{Lucicetal,BigGAN,mescheder2018training}. A lot of recent research focus has been on stabilizing this process \cite{improvementofgantraining,improvementofwasserstein,LeastSquaresGAN,shrivastava,BRE}. Since the discriminator is seen as the primary driving force, its loss is defined to minimize the distance between the training distribution and the generated distribution. Originally, Jensen-Shannon divergence was used as a distance metric~\cite{GAN}. A few more stable alternatives including least squares~\cite{LeastSquaresGAN} and Wasserstein distance \cite{wasserstein} are also commonly used. Primarily, the goal is to train the generator, the discriminator is only trained to steer the loss in the right direction through the generator and is discarded once the generator has been trained.
Adversarial training has emerged as a default choice for image translations involving one-to-many (e.g., label-to-image) or many-to-many (image-to-image) mappings.
For the generator in GANs, a pixel-based loss such as L1 is used on the output of the network~\cite{pix2pix, BiLevel}. Some improved loss functions have been proposed and used in many recent GAN based frameworks, among which a perceptual-loss~\cite{johnson2016perceptual} (computed over features of a pretrained CNN model) have been found very effective. However, as noted in~\cite{cyclegan, pix2pix} these losses alone are not sufficient and are used as added assistance with the adversarial training.

In this paper, we propose a new loss formulation termed \textit{the Spatial Profile Loss} which is measured in a way that it approximates measuring the distance between the two distributions while still being based on the pixel-based comparisons. We compute this on image gradients' space and on converted colour spaces during training and show that this has a net effect of distilling shape/content from colour/style. The proposed loss is meaningful enough to directly drive the generator for one-to-many or many-to-many image translation problems. We discuss some related work in the next section. In Section~\ref{sec:GON} we describe our proposal in detail along with its applications on diverse image translation problems. 

\vspace{-12pt}
\section{Related Work}
\label{sec:relwork}
\vspace{-4pt}

\noindent\textbf{Image-to-image translation:}
Image-to-image translation describes a broad category of problems that transfer an input image from one domain to another.
In comparison to previous feed-forward approaches that used regression terms as reconstruction losses~\cite{cheng2015deep,pathak2016context}, adversarial approaches improve the representation of frequency structure and replaced these direct feed-forward methods~\cite{pix2pixhd,karacan2016learning,SRGAN,wang2018perceptual,wang2017discriminative}.   
Isola~\etal~\cite{pix2pix} propose a generic framework using image-conditional GANs for general image-to-image translation such as label-to-photo mappings. In their approach, they combine adversarial losses with an L1 reconstruction loss, which leads to a better representation of low-frequency structures.
This basic framework was used as a stepping stone for multiple GAN-based image-to-image translation methods.
For example, Zhu~\etal~\cite{cyclegan} replace the standard L1-loss with cycle-consistency- and identity-losses by introducing another generator-discriminator pair for an unpaired setting. 
While adversarial approaches produce visually convincing results, the training process can be rather difficult and a balance between reconstruction and adversarial losses has to be found~\cite{Lucicetal}. Furthermore, as multiple networks are included, one is more restricted regarding usable resources, which in return can negatively impact the training~\cite{Lucicetal, BigGAN}.
Our work differs, as additional networks are not required, which in return opens up more possibilities regarding the network architecture and simplifies the training process.\\
Recent approaches adopt perceptual losses (either reconstruction or as adversarial) by extracting an image's high-level features of a pre-trained image classification network~\cite{chen2017photographic,pix2pixhd,wang2017discriminative,wang2018perceptual,mechrez2018contextual}. In doing so, the generators manage to create structure expected by a classification network.
For example, Chen and Koltun\cite{chen2017photographic} showed promising results on label-to-image  mapping  problem  using  cascaded refinement networks~(CRN). Here, a single generator is trained using a modified perceptual loss.
Similarly, Mechrez~\etal\cite{mechrez2018contextual} proposed a contextual loss for image transformation tasks, which represents the differences of distributions of features extracted by an external network such as a pretrained VGG19. 
 In contrast,  we propose that one can simply train a single network without the influence of such external information by investing in different loss formulations than those used at present.\\
\noindent\textbf{Single-image super-resolution:} Super-resolution has received significant attention and has been studied for decades.
Super-resolution (SR) is the process of the estimation of a high-resolution image from its low-resolution counterpart. SR methods that employ deep learning have shown significant quantitative improvement in terms of PSNR values~\cite{FSRCNN,huang,kim,EDSR}. In literature, photorealism has mostly been affiliated with GANs and now the SR problem is also being addressed with various GANs architecture in many recent works~\cite{CinCGAN, wang2018esrgan,bulat2018learn}. Recently, Ledig~\etal~\cite{SRGAN} (SRGAN) showed state-of-the-art results on SR trained with both perceptual and adversarial losses. We make use of SRGAN's generator and train it to restore SR images without adversarial and perceptual objectives. \\ 
\noindent\textbf{Makeup Transfer and removal:} Makeup transfer can be regarded as a sub-problem of style transfer, where the goal is to transfer facial makeup of a reference face image onto an input face.
Liu~\etal~\cite{liu2016makeup} follow the route of neural style transfer~\cite{gatys2015neural} and transfer the makeup locally on facial components. GANs now have produced photorealistic results on this problem. Most recently adapted CycleGAN variants~\cite{chang2018pairedcyclegan,li2018beautygan} were used for both makeup transfer and makeup removal. While Li~\etal~\cite{li2018beautygan} use a symmetric approach using a segmentation level makeup loss as an additional perceptual loss, Chang~\etal~\cite{chang2018pairedcyclegan} approach this problem in an asymmetric manner thus using distinctive local generator-discriminator pairs for both makeup transfer and removal. We show how we can directly train a standard generator for this task as well, producing photo-realistic results without the need of such an extensive adversarial training.
\vspace{-12pt}
\section{Image Translations with the Spatial Profile Loss}
\label{sec:GON}
\vspace{-4pt}
For image translation problems, it is important to preserve high frequency details while transferring low frequency content (e.g., uniform textures, colour, tone etc.,) as much as possible. The purpose of the generator is to produce an image as close as possible to the target. While in the typical GANs this is primarily achieved using the discriminator loss, a reconstruction term is also added at the generator output to minimize the difference between the target and generated image. The reconstruction loss is simply the pixel-based L2- or  L1-loss. Since this is the average of pixel-wise differences, it is not that effective in 
reconstructing  
the actual high frequency content and the low frequency details in the image.

\begin{figure}[t]
\centering
\includegraphics[width=0.9\linewidth]{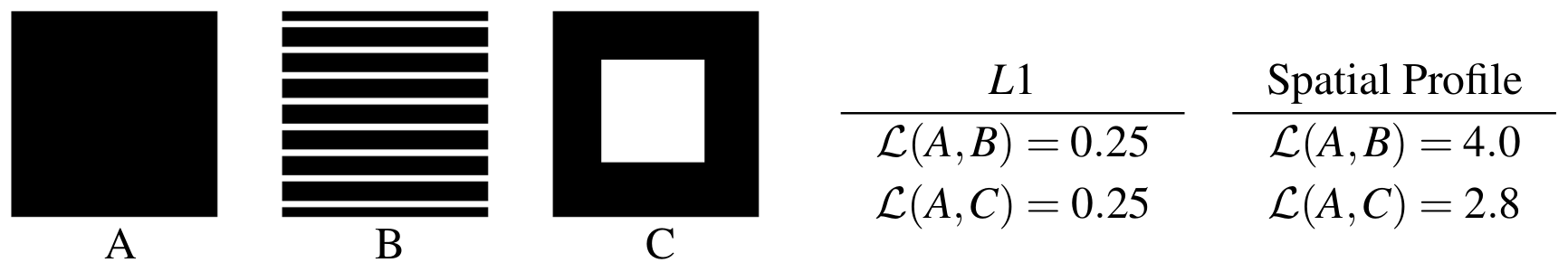}
\caption{Patterns B \& C are very different but their L1 difference with A is the same.}
\label{fig:toy_exp}
\end{figure}

To illustrate this point, Figure~\ref{fig:toy_exp} shows two very different patterns of size $64\times64$ (B,C), which have the same number of white pixels. Average pixel difference such as computed in L1-loss may not be of much information about the structural changes if the underlying average differences are about the same. If we compute a mean absolute difference between the source A and each of these targets, we get the same value of $\mathit{0.25}$. Image content or structure can be characterized better by viewing the pixel variations along a given direction in the image. Along this direction, the changing pixel values define a particular pattern that describes the underlying pixel relations.  In our example, if we consider each image column as a vector and compute the Euclidean distance between corresponding column vectors in two images, the average of these distances comes out to be $\mathit{4}$ between A \& B and $\mathit{2.8}$ between A \& C. One can see that such a matching on \textit{spatial profiles} can capture better structural and perceptual content in the image while being more informative.
\setlength{\abovedisplayskip}{4.5pt}
\setlength{\belowdisplayskip}{4.5pt}
\setlength{\abovecaptionskip}{2pt}
\setlength{\belowcaptionskip}{-14pt}

We propose a novel loss formulation around these ideas. In principle, we compute the loss between a source and target image by considering such pattern differences along the image x and y-directions. Considering a row or a column spatial profile of an image as a vector, we can compute the similarity between them in this induced vector space. Formally, this similarity is measured over each image channel 'c' as:
\begin{equation}\label{eq:1}
\begin{split}
    \mathcal{L}(x,y|G(x;\theta), y) = \sum_c & \Big(\frac{1}{H}\mathrm{trace}\hspace{1pt}(G(x;\theta)_{c} \cdot y_{c}^{\tau}) +  \frac{1}{W}\mathrm{trace}\hspace{1pt}(G(x;\theta)_{c}^{\tau} \cdot y_{c})
    \Big)
\end{split}
\end{equation}
where $(\cdot)^{\tau}$ represents transpose. The first term computes similarity among row profiles and the second among column profiles of an image pair $(x,y)$ of size $H \times W$. These image pixels profiles are L2-normalized to have a normalized cosine similarity loss. Given the proposed general scheme in Equation~\ref{eq:1}, we propose to define two loss components to drive the content/shape and style/colour of the images separately. Since an image shape is measured by its edge content, we can compute this loss with Equation~\ref{eq:1} in the image gradients space, termed Gradient Profile ($GP$) loss:
\begin{equation}\label{eq:grad}
GP(x,y)= \mathcal{L}(x,y|\boldsymbol{\nabla}G(x;\theta), \boldsymbol{\nabla}y)
\end{equation}

The image gradients in each channel can easily be computed by simple 1-pixel shifted image differences from itself. We observe in our extensive ablation studies and trainings that this $GP$ loss is quite powerful in driving gradients with healthy magnitudes resulting in fast model training in fewer epochs. We will show that it effectively distils the shape content from the colour content of the image. Similar to this, the colour content of the image can be measured e.g., by directly using Equation~\ref{eq:1} on RGB pixels. For problems such as a one-to-many mapping of colours, RGB tends to collapse to an average colour value. This has been studied well in image colourization problems~\cite{zhang2016colorful} where a different colour space such as LAB is used. In contrast, we observed that training the network on converted colour space images is not that effective and propose to only compute the loss in a different colour space. For this, we use the luma-chrominance YUV conversion. On this converted space the chrominance UV channels carry the colour information and computing the loss in this space transfers the colours effectively. Additionally, computing the loss in the gradient space of the chrominance channels helps the network learn stronger colour transfer and it preserves fine details driven by the colour gradients. We show on typical one-to-many mapping problems that it effectively learns the colour mappings. Our colour loss, therefore is the sum of the three terms, the RGB, YUV and $\boldsymbol{\nabla}$YUV loss. We define our Colour Profile ($CP$) loss as:
\begin{equation}\label{eq:YUV}
\begin{split}
CP(x,y)= &\mathcal{L}(G(x;\theta)^{\scaleto{RGB}{3pt}}, y^{\scaleto{RGB}{3pt}}) + \mathcal{L}(G(x;\theta)^{\scaleto{YUV}{3pt}}, y^{\scaleto{YUV}{3pt}}) + \mathcal{L}(\boldsymbol{\nabla}G(x;\theta)^{\scaleto{YUV}{3pt}}, \boldsymbol{\nabla}y^{\scaleto{YUV}{3pt}})
\end{split}
\end{equation}

The combined loss to drive the generator network, termed as Spatial Profile Loss (SPL) is the sum of $GP$ and $CP$ losses. 
Note that since Equation~\ref{eq:1} computes a similarity, we can minimize the negative of this as:
\begin{equation}\label{eq:GON}
\mathcal{L}_{SPL}(x,y)=\argmin_G -(GP(x,y) + CP(x,y))
\end{equation}
We show in our experiments that these two terms distils the shape and colour/tone of the image and provide control in driving the two terms with different targets, e.g., we can compute the $GP$ loss between output and input and the colour loss between output and target to train a generator in an auto-encoder fashion for style-transfer applications.
\newcommand{\f}[1]{\textbf{#1}} 
\newcommand{\np}[0]{$^\dag$}  
\newcommand{\tnote}[1]{\tiny{\textcolor{red}{#1}}}

\newsavebox\zbox
\newcounter{zoom}

\newcommand{\zoombox}[2][1]{
  \leavevmode
  \sbox\zbox{#2}%
  \pdfdest name {zb\thezoom.in} fitr
    width  \wd\zbox\space
    height \ht\zbox\space
    depth  \dp\zbox\space
  \makebox[0pt][l]{\fboxsep=0.5\fboxsep\hskip-\fboxsep%
    \immediate\pdfannot
      width  \dimexpr\wd\zbox+2\fboxsep\relax\space
      height \dimexpr\ht\zbox+\fboxsep\relax\space
      depth  \dimexpr\dp\zbox+\fboxsep\relax\space
    {
      /Subtype/Link/Border [0 0 #1[#1]]
    }%
    \immediate\pdfannot
      width  \dimexpr\wd\zbox+2\fboxsep\relax\space
      height \dimexpr\ht\zbox+\fboxsep\relax\space
      depth  \dimexpr\dp\zbox+\fboxsep\relax\space
    {
      /Subtype/Widget/FT/Btn/Ff 65536/H/N
      /T (zb\thezoom)/TU (Press mouse button to zoom in.)
      /AA <<
        /D <</S/GoTo /D (zb\thezoom.in)>>
        /U <</S/Named /N/GoBack>>
      >>
    }%
  }%
  \usebox{\zbox}%
  \stepcounter{zoom}%
}
\ifdefined\pdfextension
  \protected\def\pdfannot{\pdfextension annot }
  \protected\def\pdfdest{\pdfextension dest }
\fi
\vspace{-10pt}
\subsection{Image-to-Image Domain Mapping}\label{seq:i2i}
\vspace{-0pt}
\setlength{\abovecaptionskip}{2pt}
\setlength{\belowcaptionskip}{-15pt}
\begin{figure*}[t!]
\hspace{-10pt}
\centering
\includegraphics[width=1.025\linewidth]{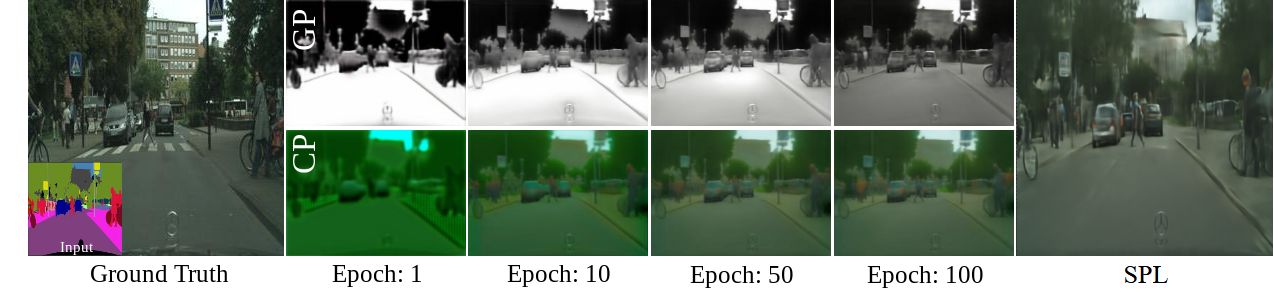}
\caption{Loss Analysis. We train separately with shape ($GP$) only and colour ($CP$) only loss to show what they drive. We display the result at evolution of epochs during training on an image from Cityscapes validation set and also show the final output with our combined $SPL$ on label2image.}
\label{fig:Cityscapes_ablation}
\end{figure*}
\setlength{\abovecaptionskip}{-0pt}
\begin{figure}[b!]
\centering
\includegraphics[width=\linewidth]{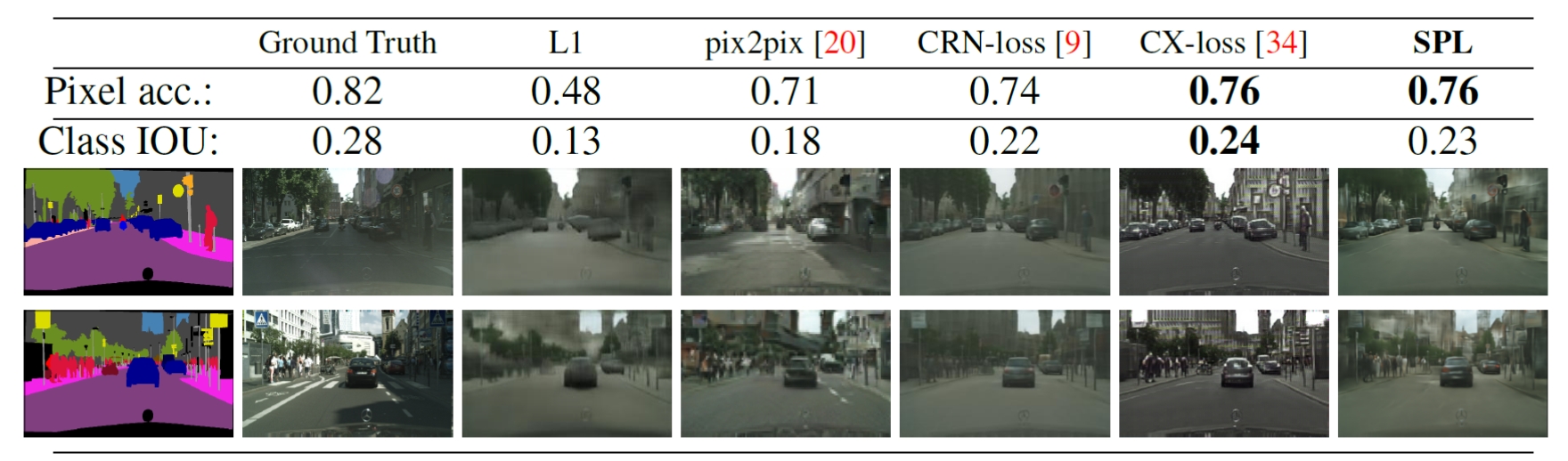}
\caption{Loss Comparison: Same architecture trained on the Cityscapes label2image with different loss functions. \textbf{Top}: Averaged FCN-8-scores are provided on the full validation set.}
\label{tab:citscapes}
\end{figure}

In this section, we show the effectiveness of SPL for general image-to-image domain mappings. We take the same 9 block ResNet encoder-decoder generator network used in the pix2pix and Cyclegan frameworks of Zhu~\etal~\cite{cyclegan}. We then directly train it on 256x256 images of several domain mapping problems (night2day, the thermal2visible Carl face dataset~\cite{espinosa2013new}, edges2shoes and Cityscapes) using the proposed $\mathcal{L}_{SPL}$ loss. We use a minibatch size of 16 with a fixed learning rate of $2\times 10^{-4}$ for 50, 50, 150 and 150 epochs respectively.

\noindent\textbf{Experiments:} In Figure~\ref{fig:Cityscapes_ablation}, we conduct an \textbf{ablation study} in order to demonstrate the effect of $GP$ and $CP$ loss component and how it distills and drive the shape and colour respectively. The figure shows the results for SPL trained with only $GP$ or only $CP$ at 1, 10, 50 and 100 epochs from left to right,  making it apparent that $GP$ enforces the structure of the target image while $CP$ influences the colour transfer. The final combined effect of $GP$ and $CP$ using the $\mathcal{L}_{SPL}$ is presented at the end which shows that $CP$ loss transfers the colour properties to the shape enforced by $GP$ to create the synthesized image.

 Figure~\ref{tab:citscapes} provides comparison of different loss functions on the Cityscapes label2image task with quantitative comparisons at the top and qualitative results in the bottom. We train the pix2pix generator with only L1 loss, Pix2Pix's losses, and state-of-the-art feature-level perceptual losses i.e., CRN~\cite{chen2017photographic} and Contextual (CX)~\cite{mechrez2018contextual} loss formulations. The performance is compared on the commonly used averaged FCN8-scores\cite{long2015fully, pix2pix} on the full validation set. 
 The perceptual feature-level losses CX and CRN are computed by using all of the features maps of the pre-trained VGG19 network as it produces the best results for these losses. We have used the respective author's official loss implementations.
In terms of metrics, the proposed SPL formulation not only outperforms its directly comparable pixel-level L1-loss but is also on par with the adversarial (GAN) and perceptual feature level losses.

When comparing the results qualitatively in Figure~\ref{tab:citscapes}, our SPL formulation presents a sensible distribution of colours on different classes. Pedestrians and cars are shown in differing separated colours. Similarly, the model maps reasonable outputs onto traffic signs. The feature level losses learn distinct shapes and sharper details but have a more generalized output for each class without much variability in the textures or colour details on the objects.
Nevertheless, SPL has shortcomings when it comes to bringing out sharper structure on background e.g., buildings, as it defaults to a general blurry shape. This could be because the details on the buildings are not consistent and have much more variability to learn than is achievable by computing the loss over vertical or horizontal pixel profiles. The problem could be approached, for example, by exploring other spatial profile paths that may capture this information better. 
Overall, SPL is able to learn one-to-many mappings while only using pixel-level information and as such not needing assisting networks \textendash{ discriminators and/or perceptual-networks} \textendash{ to train.}
  \setlength{\abovecaptionskip}{0pt}
\setlength{\belowcaptionskip}{-15pt}
\begin{figure*}[t]
\centering
\includegraphics[width=\linewidth]{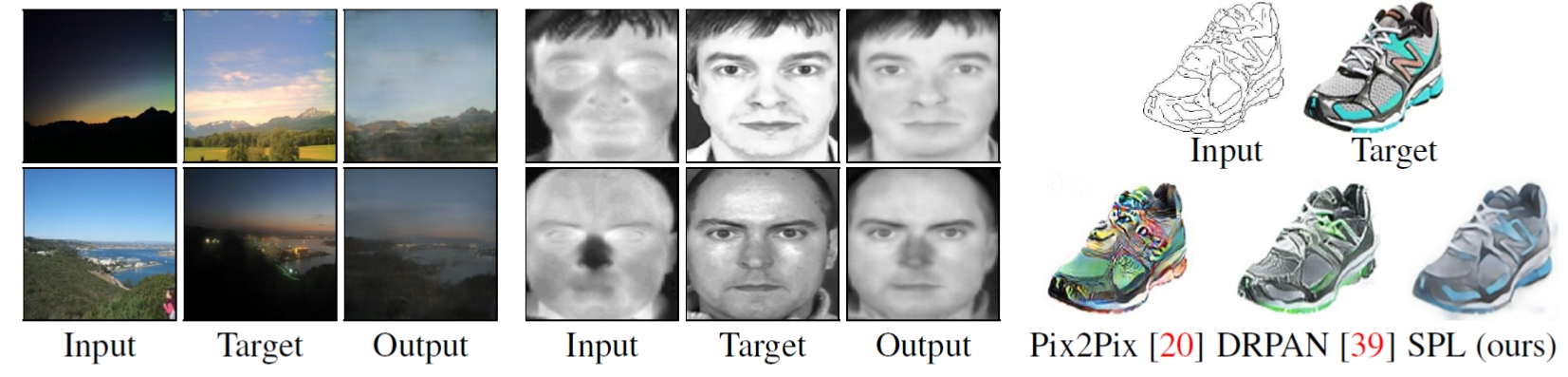}
\caption{SPL on various domain mapping tasks: Day-night (Left), thermal-visible (center) and edges2shoes (Right).}
\label{fig:style_transfer}
\end{figure*}

Figure \ref{fig:style_transfer} shows qualitative results of our trained SPL model on the night2day(top row) and day2night transformation(bottom row). In the middle column, we show the results of the mapping of thermal images to visible face images. It's quite clear that our models learn an identity-preserving mapping for each paired set of thermal and visible facial images. We provide details of this training in the supplementary. The last column shows a qualitative comparison with 
Isola~\etal~\cite{pix2pix} and Wang~\etal~\cite{wang2017discriminative} on a sample of the edges2shoes test data. In direct comparison, the same generator trained with adversarial loss~\cite{pix2pix} has problems when mapping colour on complex input structure, while SPL creates a highly plausible colour mapping as the result stays inside the boundaries of edges representing shoelaces etc.

\subsection{Single Image Super Resolution}\label{seq:resolution}
\vspace{-4pt}

\setlength{\abovecaptionskip}{5pt}
\setlength{\belowcaptionskip}{-5pt}
\begin{figure*}[t!]
\centering
\includegraphics[width=\linewidth]{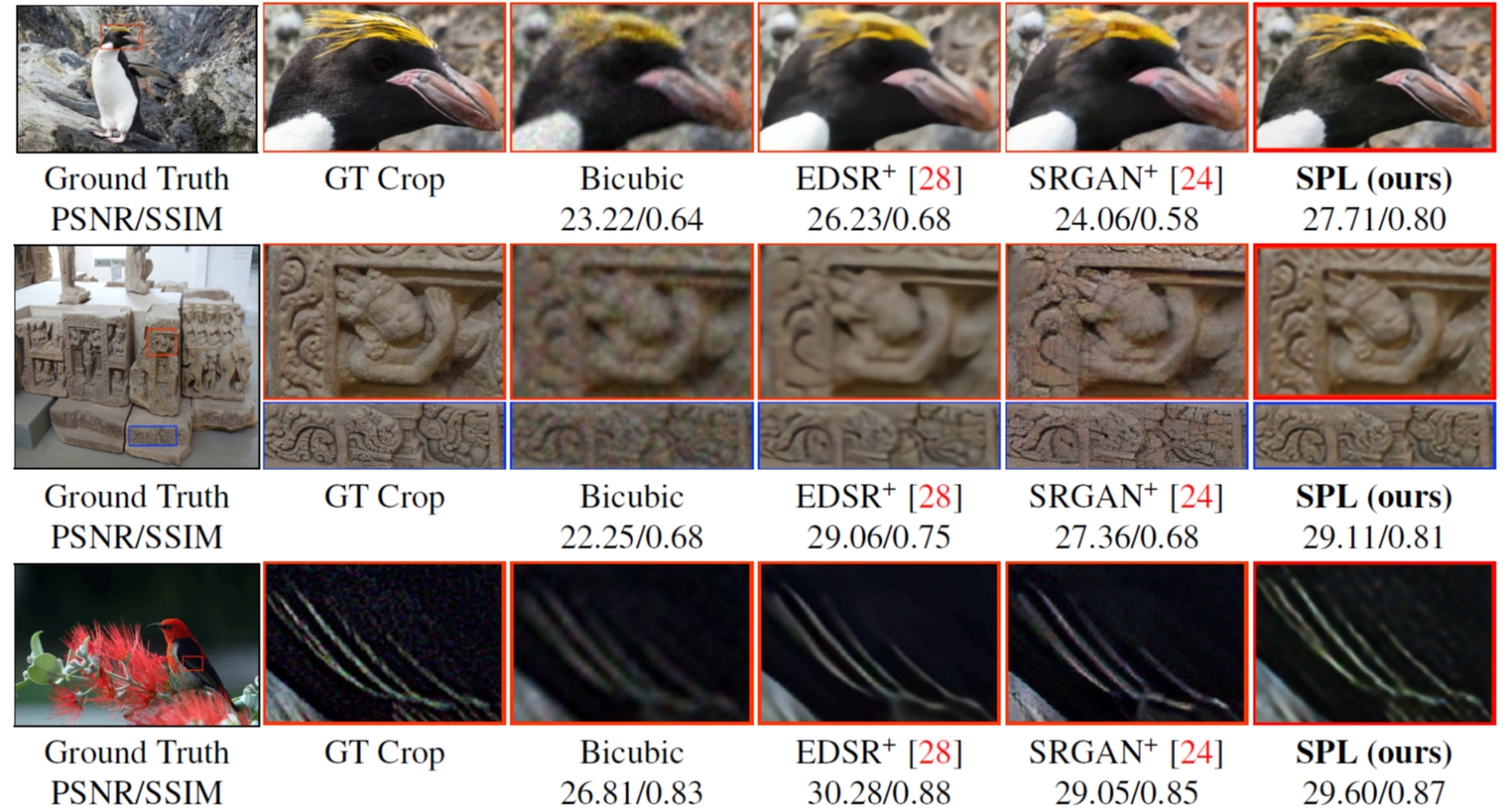}
\caption{Super-resolution results of '0801', '0816' and '0853' (DIV2K) with scale factor $\times 4$.  Full resolution ground truth (left column) with output of different methods shown. SPL (ours) achieves the state-of-the-art and restore sharper details.}
\label{fig:SRSPL}
\end{figure*}
\setlength{\abovecaptionskip}{2pt}
\setlength{\belowcaptionskip}{-9pt}
\begin{table*}[t!]
\small
    \centering
     \begin{tabular}{|c | c | c | c | c | c  | c|}
        \hline
        \footnotesize Method & \footnotesize bicubic & \footnotesize FSRCNN~\cite{FSRCNN} & \footnotesize EDSR~\cite{EDSR} &  \footnotesize EDSR\textsuperscript{+}~\cite{CinCGAN} & \footnotesize SRGAN\textsuperscript{+}~\cite{SRGAN} & \footnotesize \textbf{SPL (ours)}  \\ [0.5ex] 
        \hline\hline
        \footnotesize PSNR & 22.85 & 22.79 & 22.67 & 25.77 & 24.33 & \textbf{26.03}\\ 
        \hline
        \footnotesize SSIM & 0.65 & 0.61 & 0.62 & 0.71 & 0.67 & \textbf{0.75}\\[1ex] 
        \hline
    \end{tabular}
     \caption{Quantitative evaluation on DIV2K track 2 dataset of the proposed SPL, in terms of PSNR and SSIM.}
\label{table:PSNR_SSIM}
\end{table*}

\setlength{\abovecaptionskip}{5pt}
\setlength{\belowcaptionskip}{-13pt}
Single Image Super-Resolution (SISR)is an important image translation problem being addressed with GANs~\cite{SRGAN}.
We train the generator architecture used in SRGAN~\cite{SRGAN} and EDSR~\cite{EDSR} directly with our SPL loss on 4 times unknown downscaled LR images. For training, we take a random patch of 96x96 from the input images to produce HR resolution image upscaled by a factor of 4. Note that the generator network can take input of any arbitrary size as it is a fully convolutional network.
The network is trained from scratch for 50 epochs as compared to its SRGAN counterpart which was trained for 250 epochs.\\  
\noindent\textbf{Dataset:} We use the track 2 data from DIV2K~\cite{DIV2K} dataset for training. Track 2 is highly challenging as it contains downscaled LR images with unknown degradations (noise, blurring and pixel shifting).
This dataset contains 800 training images and 100 validation images which cover a variety of natural scenarios i.e. animals, person images, buildings. As with previous methods, the performance is tested on the provided 100 validation images.
\noindent\textbf{Results and Comparison:} We compare the performance of our proposed SPL on DIV2K track 2 with state-of-the-art supervised SISR methods: FSRCNN~\cite{FSRCNN}, EDSR~\cite{EDSR} and SRGAN~\cite{SRGAN}. For quantitative evaluation, we use PSNR and structural similarity SSIM. The results for FSRCNN, EDSR and SRGAN have been taken from Yi et al.~\cite{CinCGAN}. They show improved results for EDSR and SRGAN by first training these models on track 1 data (bicubic downscaling) and then fine-tuning on track 2 data (EDSR\textsuperscript{+}, SRGAN\textsuperscript{+}). Table \ref{table:PSNR_SSIM} shows the mean PSNR and SSIM values of restored validation images. It illustrates that FSRCNN and EDSR perform worse if the blur and noise are unknown in the training process. Fine-tuning with track 2 improves the results for EDSR\textsuperscript{+} and SRGAN\textsuperscript{+}. Our model being trained from scratch with SPL on track 2 achieves the state-of-the-art in comparison to the considered SISR methods
Figure \ref{fig:SRSPL} shows a more subjective analysis of chosen validation images. For every example image, the ground truth target and restored images are displayed with cropped details in the subsequent row.  As can be seen, where EDSR\textsuperscript{+} manages to get comparable PSNR values, it restores smoother regions lacking high frequency edge information and fine structure details. In contrast, SPL has learned to distill high frequency details through the $GP$ loss and restored sharper colour content via the $CP$ component of the proposed $\mathcal{L}_{SPL}$. We include the full SR output of these methods in the supplementary material.

\vspace{-4pt}
\subsection{Photo Realistic Makeup Transfer}\label{seq:makeup}
\vspace{-2pt}
\setlength{\abovecaptionskip}{3pt}
\setlength{\belowcaptionskip}{-10pt}
\hspace{50pt}
\begin{figure}[t]
    \centering
    
\includegraphics[width=\linewidth]{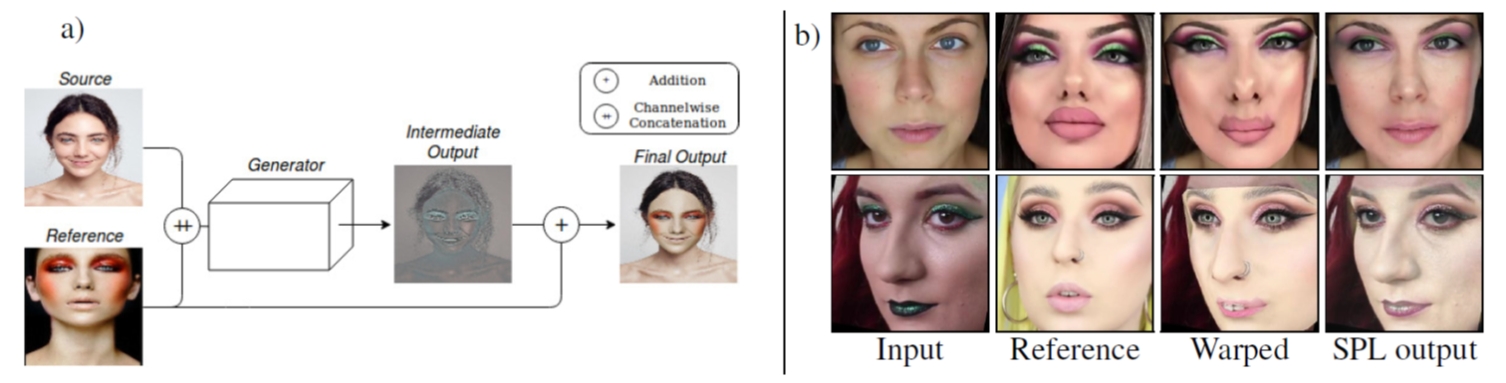}
    \caption{a). Proposed makeup transfer sequence: Source and reference image are passed to a generator. Its result is added onto the reference, resulting in the final output, over which the $\mathcal{L}_{SPL}$ loss is computed. b). Preprocessing through warping and corresponding results. }
    \label{fig:makeup_diagram}
\end{figure}

We demonstrate the usefulness of distilling shape from colour and the control we have in our loss formulation on makeup transfer applications. 
We propose a novel and effective design for training a single network for such problems, see Figure \ref{fig:makeup_diagram}(a). Here, we combine the source and the reference image to a 6-channel input. Our generator creates an intermediate result, which is added to the reference to reach the final output. 
We use our $GP$ loss between the generated output and the source image while the $CP$ loss component is computed between generator output and reference image. This way the network can easily reproduce the source face with all its structure preserved while only transferring makeup from the reference and allows us to train on unpaired data.
As this approach is not restricted to a specific network, we can adjust it to any problem at hand.

We use the generator architecture of CycleGAN~\cite{cyclegan} for standard and the SR generator (used in section~\ref{seq:resolution}) for $\times 4$ SR makeup transfer tasks. 
For our training, we warp the reference face on the source image as shown in Figure \ref{fig:makeup_diagram}(b).  The warping process can lead to different types of artefacts, however, our colour transfer is mostly not affected by this. Each model is trained for 10 epochs with a batch size of 10.

\noindent\textbf{Proposed FCC Dataset:} 
  Existing makeup transfer databases are relatively small\cite{FAM,YMU} (less than 1000 images each) or were, to our knowledge at the time, not publicly available. We, therefore, constructed a new \textbf{F}acial \textbf{C}osmetic \textbf{C}ontent database (FCC) for strong makeup as well as general makeup in both higher and lower resolutions.
We detected face images from Youtube makeup tutorials from various regions depicting racial diversity. Images from the first and last quarter of each video are kept in order to capture faces with and without makeup.
We then classify the extracted images as strongly visible makeup, general makeup, non-makeup by extracting CNN features~\cite{VGG2}. After removing remaining noise, 
we re-sized the images to either $256\times256$ or $512\times512$.
This way, we gather 5389 non-makeup and 13036 makeup images out of which 5281 contain strongly visible makeup. The FCC dataset with a total of $\sim 18400$ images is currently the largest dataset for makeup style transfer applications.
Sample images of the dataset and more details are provided in the supplementary.

\begin{figure*}[t]
\vspace{1pt}
\centering
\includegraphics[width=\linewidth]{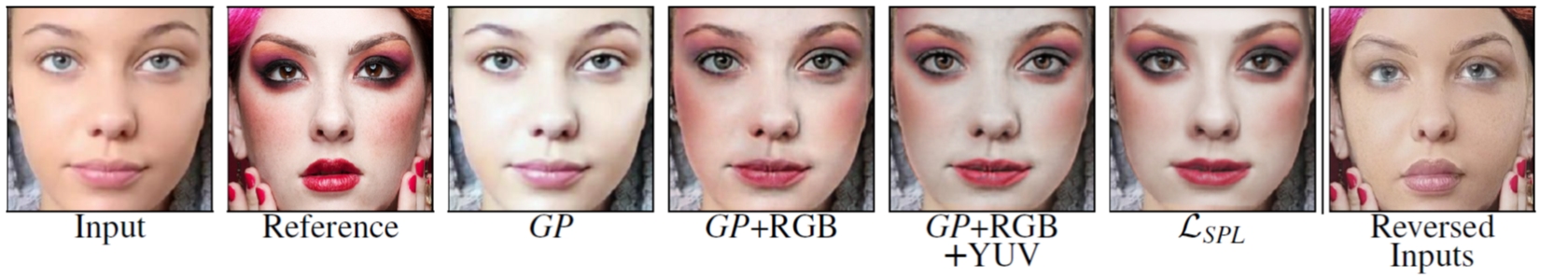}
\caption{Loss Analysis. Comparison of impact of different loss terms for makeup transfer. From left to right: Input, reference, $GP$, $GP$+RGB-profile loss,GP+RGB+YUV-profile loss, full $\mathcal{L}_{SPL}$ as well as $\mathcal{L}_{SPL}$ with reversed inputs.}
\label{fig:ablation_makeup}
\end{figure*}

\noindent\textbf{Experiments:} 
 In Figure \ref{fig:ablation_makeup}, we perform an \textbf{ablation study} to analyse the influence of each loss term by separately training the generator each time with one part of the described losses added with equal weight to the total loss. 
Considering only $GP$ between source and output, the generator resurfaces the underlying structure of the source image, while removing most of its colour. After including the RGB profile loss between source and reference, the generator learns to transfer the basic colour scheme of the reference image, without being impacted by portrayed facial features like eye wrinkles or freckles. However, it cannot translate visual details as it is not able to grasp brightness variations properly. The YUV-profile loss term lets the generator recreate better tone transitions.
By adding the YUV-gradient loss term ($\boldsymbol{\nabla}$YUV), the model learns to better represent brightness transitions, thus improving the saturation and leading to a more natural transfer. Note that because of our proposed design, interestingly, the same model trained for makeup transfer 
also manages to remove makeup when swapping source and reference, as depicted in the last snippet of Figure~\ref{fig:ablation_makeup}.
\setlength{\abovecaptionskip}{-5pt}
\setlength{\belowcaptionskip}{-15pt}

\setlength{\abovecaptionskip}{0pt}
\setlength{\belowcaptionskip}{-15pt}
\begin{figure*}[b!]
\centering
\vspace{-2pt}
\includegraphics[width=1.01\linewidth]{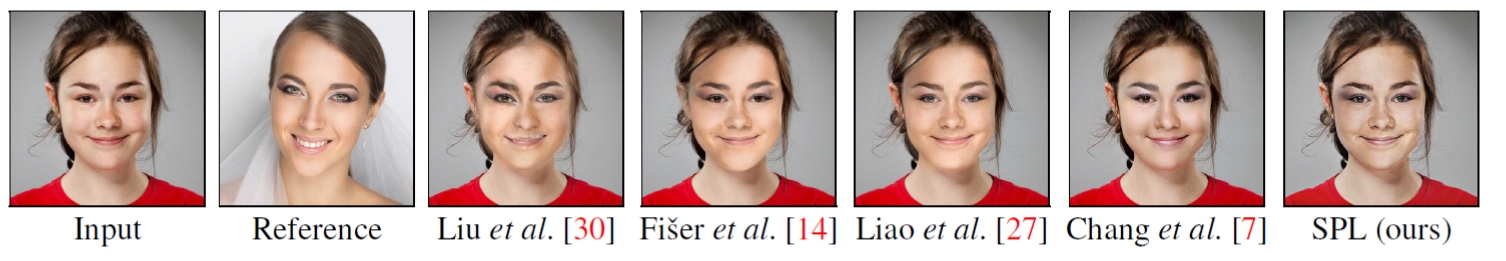}
\caption{State-of-the-art comparison between different makeup transfer methods. 
}
\label{fig:makeup_comparison}
\end{figure*}

\begin{figure}[t!]
\centering
\vspace{-0pt}
\includegraphics[width=\linewidth]{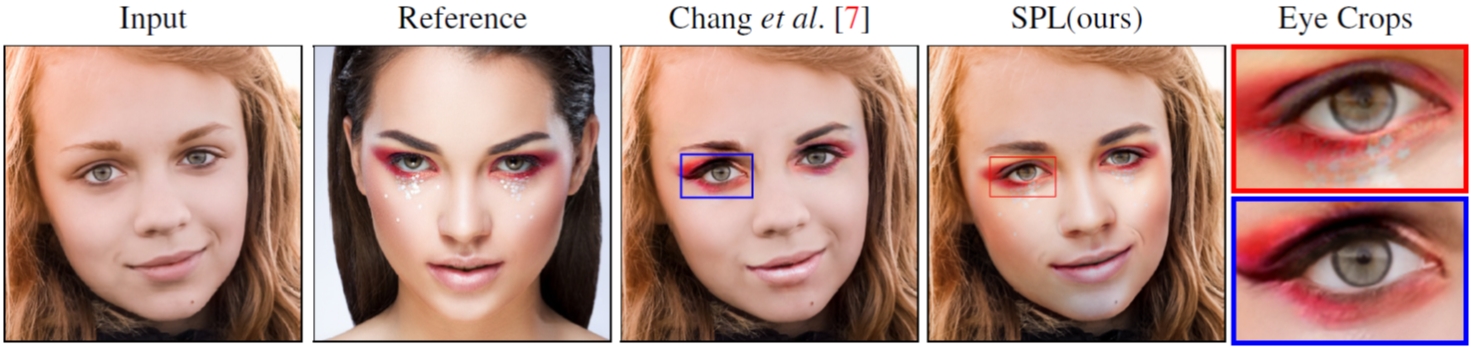}
\caption{Comparison with Chang~\etal\cite{chang2018pairedcyclegan} on extreme makeup styles. We demonstrate our SR-SPL model for makeup transfer. Eye closeup: Chang~\etal (bottom) and ours (top). 
}
\label{fig:hires_makeup}
\end{figure}

\noindent\textbf{Comparison with state-of-the-art:} In  Figure  \ref{fig:makeup_comparison},  we compare our results to four different previous works.
Liu~\etal~\cite{liu2016makeup} appears to be affected by factors such as pose and expression variations that can lead to artefacts like asymmetric eyes or overlaps between source and reference.
 Fi\v{s}er~\etal's~\cite{fivser2017example} approach applied to makeup transfer can show problematic tone transitions between skin regions.
Liao~\etal~\cite{liao2017visual} 
works well on areas sharing similar features while the resulting colour transfer suffers in the case of excessive makeup regions like eyeshadows.
Most recent work by Chang~\etal~\cite{chang2018pairedcyclegan} train a CycleGAN variant each on eye, mouth and skin regions as the makeup between these regions can strongly vary. While this approach achieves makeup of quality comparable to the reference, the generated image might not display a correct transfer from it, as it will only recreate styles already seen in training. 
In contrast,
our network learns to preserve the source's facial attributes such as expression and pose and only transfer the reference's makeup. 
Since our model only learns a colour transfer between source and reference image, we are robust to styles different to our training data and so can replicate any form of makeup ranging from only affecting small areas to full-face makeups. 
It is sufficient to train our single network for only 10 epochs with SPL, while Chang~\etal~\cite{chang2018pairedcyclegan}  train a multitude of generator-discriminator pairs for at least 400 epochs.
In Figure \ref{fig:hires_makeup}, we compare with Chang~\etal~\cite{chang2018pairedcyclegan} on an extreme makeup style that was unseen during either training. 
For this experiment, we show the result of our SR makeup transfer producing full HD $2048\times2048$ images for LR $512\times512$ source and reference images (see also Figure~\ref{fig:intro}). 
While the result by Chang~\etal~\cite{chang2018pairedcyclegan} seems reasonable, it fails to replicate fine colour details shown by the reference.
SPL fully captures the makeup tone on the cheeks, sparkling makeup around the bottom eyelid as well as the star-shaped structures beneath it.

We include more details and the limitations of our method in the supplementary material.  Our code along with the proposed FCC dataset is available online\footnote[1]{\url{https://github.com/ssarfraz/SPL}}.

\setlength{\abovecaptionskip}{1pt}
\setlength{\belowcaptionskip}{-10pt}

\vspace{-4pt}
\section{Conclusion}
\label{sec:conclusion}
\vspace{-4pt}
The Spatial Profile Loss formulation makes use of both gradients and chromatic channels to approximate distances in shape and color between two images. A unique property of the proposed formulation is the use of loss calculations on pixel patterns defined by horizontal or vertical profiles. The central idea is to approximate the image distribution in a low-dimensional subspace spanned by these pixel-profiles vectors. While we have shown that simple vertical/horizontal profiles suffice for approximating the target distribution one can explore more orientations. In fact we have experimented with including the diagonal profiles, nevertheless, with negligible qualitative improvements. As a regular sampling method using specific orientations, it's possible there are attributes of an image that are not being captured. For future work, including more paths such as random walks or other types of paths would be interesting, if not as efficient to compute.

Conclusively, we have presented an effective loss formulation based on spatial profiles for general image translation problems. 
The proposed SPL formulation enables us to independently drive the shape and colour which, among others, is useful in typical style/colour transfer problems. On diverse image translation tasks, our results demonstrate that the proposed loss components can effectively minimize the difference between the generated and the target image distributions. This suggests that even without adversarial training or additional perceptual networks realistic transfer is possible and can even surpass its counterpart.  We hope that our work may suggest seeking alternatives and more effective ways to train the generators for similar problems.

{\small
\bibliographystyle{ieee}
\bibliography{egbib}
}

\end{document}